\documentclass[10pt,twocolumn,letterpaper]{article}

\usepackage{cvpr}              

\usepackage{graphicx}
\usepackage{amsmath}
\usepackage{amssymb}
\usepackage{booktabs}

\usepackage{times}
\usepackage{epsfig}
\usepackage[table,dvipsnames]{xcolor}
\definecolor{myblue}{RGB}{47, 114, 173}

\usepackage{tocloft}
\usepackage{titletoc}
\newcolumntype{L}[1]{>{\raggedright\let\newline\\\arraybackslash\hspace{0pt}}m{#1}}
\newcolumntype{C}[1]{>{\centering\let\newline\\\arraybackslash\hspace{0pt}}m{#1}}
\newcolumntype{R}[1]{>{\raggedleft\let\newline\\\arraybackslash\hspace{0pt}}m{#1}}

\usepackage[accsupp]{axessibility}  

\usepackage[pagebackref,breaklinks,colorlinks]{hyperref}

\usepackage{pifont}
\newcommand{\cmark}{\ding{51}}
\newcommand{\xmark}{\ding{55}}
\definecolor{myred}{RGB}{220,50,47} 
\definecolor{mygreen}{RGB}{133,153,0}
\definecolor{commentcolor}{RGB}{133,153,0}
\definecolor{urlcolor}{rgb}{0.93,0.01,0.55}
\newcommand{\greencmark}{\textcolor{mygreen}{\cmark}}
\newcommand{\redxmark}{\textcolor{myred}{\xmark}}
\newcommand*{\affmark}[1][*]{\textsuperscript{#1}}

\usepackage{mathtools}

\DeclareMathOperator*{\argmin}{arg\!\min}  
\mathchardef\mhyphen="2D

\newcommand{\citep}{\cite}

\usepackage[capitalize]{cleveref}
\crefname{section}{Sec.}{Secs.}
\Crefname{section}{Section}{Sections}
\Crefname{table}{Table}{Tables}
\crefname{table}{Tab.}{Tabs.}

\def\cvprPaperID{27} 
\def\confName{CVPR}
\def\confYear{2023}

\begin{document}

\title{On Assimilating Learned Views in Contrastive Learning}
\title{Constructive Assimilation: Boosting Contrastive Learning Performance \\through View Generation Strategies}

\author{Ligong Han\affmark[1,2]\thanks{Work done during an internship at MIT-IBM Watson AI Lab.}\quad Seungwook Han\affmark[2]\quad Shivchander Sudalairaj\affmark[2] \quad Charlotte Loh\affmark[3]\\Rumen Dangovski\affmark[3]\quad Fei Deng\affmark[1] \quad Pulkit Agrawal\affmark[4] \quad Dimitris Metaxas\affmark[1]\\ Leonid Karlinsky\affmark[2] \quad Tsui-Wei Weng\affmark[5]\quad Akash Srivastava\affmark[2]\\
{\affmark[1]Rutgers University\quad\affmark[2]MIT-IBM Watson AI Lab\quad\affmark[3]MIT EECS\quad\affmark[4]MIT CSAIL\quad\affmark[5]UCSD}
}

\maketitle

\begin{abstract}
   Transformations based on domain expertise (expert transformations), such as \textit{random-resized-crop} and \textit{color-jitter}, have proven critical to the success of contrastive learning techniques such as SimCLR. Recently, several attempts have been made to replace such domain-specific, human-designed transformations with generated views that are learned. However for imagery data, so far none of these view generation methods has been able to outperform expert transformations. In this work, we tackle a different question: instead of replacing expert transformations with generated views, can we constructively assimilate generated views with expert transformations? We answer this question in the affirmative and propose a view generation method and a simple, effective assimilation method that together improve the state-of-the-art by up to $\approx 3.6\%$ on three different datasets. Importantly, we conduct a detailed empirical study that systematically analyzes a range of view generation and assimilation methods and provides a holistic picture of the efficacy of learned views in contrastive representation learning.
\end{abstract}

\section{Introduction}
\begin{figure}[t]
    \centering
    \includegraphics[width=1\linewidth]{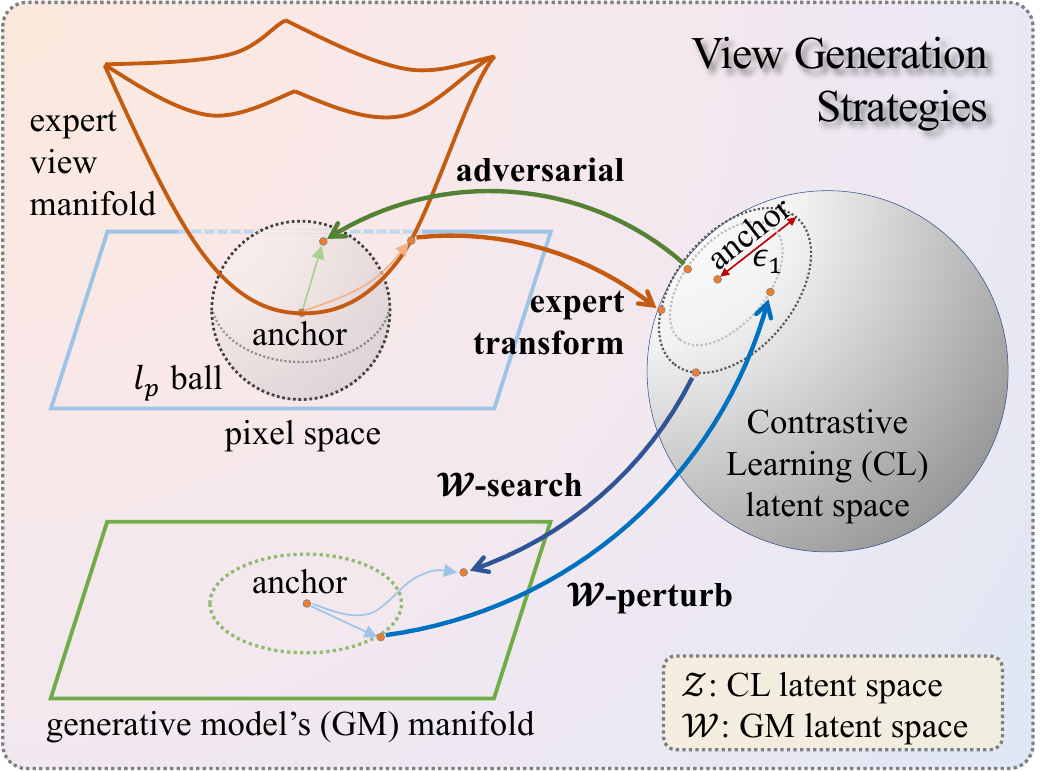}
    \caption{Illustrative visualization of the view generation strategies. The convex shaped surface represents the expert view manifold, and the upper blue and lower green planes represent the pixel space and the GAN's manifold, respectively. The hypersphere represents the learned representation space of a CL encoder. \textbf{$\mathcal{W}$\textit{-search}} searches in the CL model's latent space ($\mathcal{Z}$-space) using the loss in \cref{eq:gan} by optimizing $w$. \textbf{$\mathcal{W}$\textit{-perturb}} directly perturbs the GAN's latent space ($\mathcal{W}$-space) with Gaussian noise. \textbf{Adversarial} (or ViewMaker~\cite{tamkin2020viewmaker}) performs adversarial training and constrains the generated view in an $l_p$ ball around the anchor in pixel space. \textbf{Expert transform} generates views by applying a set of transformations commonly used in CL.}
    \label{fig:gan}
\end{figure}
Contrastive learning (CL) has become one of the most powerful tools for self-supervised learning. Most contrastive methods are trained using instance discrimination: pulling positive views (generated from the same image) close in the learned representation space, while pushing negative views (generated from other images) away \cite{dosovitskiy2014discriminative,wu2018unsupervised,DBLP:journals/corr/abs-2002-05709,chen2020big,he2020momentum,chen2020improved,caron2020unsupervised,grill2020bootstrap,chen2021exploring, caron2021emerging,kotar2021contrasting,tejankar2021isd,wang2021solving,zbontar2021barlow}. Intuitively, the quality of representations learned by CL is highly dependent on the mechanisms for generating these views, which are commonly compositions of a set of handcrafted transformations designed by human experts. The choice of the transformations and the specific design of their composition relies on domain expertise built by years of CL research, which mostly focused on general imagery, commonly captured with consumer cameras and collected from the web. However, this domain expertise may not necessarily hold for new unseen visual domains that could be encountered by CL practitioners. Therefore, a natural question arises: {\em How can we go beyond these expert-designed views?} 

A recent theoretical analysis of CL \cite{zimmermann2021contrastive} states that CL inverts the true data generative process. This motivates us to explore the use of a generative model (a GAN in this paper) as a view generator for training a contrastive encoder. There are previous works that study view generation through the lens of {\em generative models}. Viewmaker~\cite{tamkin2020viewmaker} employs an image-to-image network and generates views via adversarial training. The generated perturbation is bounded in a $l_p$ ball in the pixel space, limiting its applicability to CL that commonly requires strong augmentation for best results. 
Recently, \cite{jahanian2021generative} defines views via a black-box generative model by perturbing slightly in the generative model's latent space without explicitly controlling the effect of this perturbation in the contrastive encoder latent space.
According to analysis in \cite{zimmermann2021contrastive}, in the ideal case, the generative and the contrastive encoder's latent spaces should be related by an orthonormal transformation, making transformations in the GAN's latent space and in the contrastive encoder's latent space equivalent up to rotation. Moreover, they empirically show that such an equivalence holds to some extent when the theoretical conditions are not satisfied.

\section{Background and Related Work}
SimCLR \cite{DBLP:journals/corr/abs-2002-05709,chen2020big} is one of the first and most established CL baselines.
Let $x \in \mathcal{X}$ denote an image {in a mini-batch $\mathcal{B}=\{x_i\}_{i=1}^N$}. Further, let $f: \mathcal{X} \mapsto \mathcal{W}$ be a representation encoder, parameterized as a deep neural network. Here $\mathcal{X} \subseteq \mathbb{R}^D, \mathcal{W} \subseteq \mathbb{R}^K$ such that $K < D$. Following the notations from \cite{khosla2020supervised}, we use $i \in \mathcal{I} \equiv \{1, \ldots, 2N\}$ to denote the index of an arbitrary batch (augmented using \textit{expert transformations} as explained below), where $j(i)$ is the index of the other augmented sample originating from the same data sample. $\mathcal{A}(i) := \mathcal{I} \backslash \{i\}$ is the complement of $i$. The contrastive loss function for SimCLR is the InfoNCE loss, which is
\begin{align}
    L_\text{simclr} = -\sum_{i \in \mathcal{I}}{\log{ \frac{\exp(z_i \cdot z_{j(i)}/\tau)}{\sum_{a \in \mathcal{A}(i)} \exp(z_i \cdot z_{a}/\tau)} }}\label{eq:simclr}
\end{align}
\noindent where, $z=f(x)$ and $\tau$ is the temperature parameter.

\paragraph{Views}
In contrastive learning, a stochastic data augmentation module applies two different sets of transformations to generate two correlated views of the same data point, $x_i$ and $x_{j(i)}$. The two sets of transformations, $t_1$ and $t_2$, are sampled from the same family of transformations $ \mathcal{T}$, such as cropping and resizing, horizontal flipping, color distortion, etc. The images $x_i$ and $x_{j(i)}$ are the positive views, whereas all other views generated from other samples are the negative views. As studied by many recent works ~\cite{Ye2019UnsupervisedEL, Misra2020SelfSupervisedLO, Tian2020ContrastiveMC}, $\mathcal{T}$ defines the invariances that the model learns. Therefore, human domain knowledge and extensive years of research have been invested to search for the optimal composition of augmentations and their corresponding parameters to optimize CL methods for a given data. We refer to the optimal set of transformations in SimCLR as \textit{expert transformations} and views created by them as \textit{expert views}.

\paragraph{View Generation }\label{sec:gan}
Previous works have explored the idea of generating views for SSL. ADIOS~\cite{shi2022adversarial} and Viewmaker~\cite{tamkin2020viewmaker} both study adversarial methods. ADIOS learns a masking function and an image encoder performing a min-max optimization on the same objective function in the masked image model framework. Alternatively, Viewmaker learns a bounded perturbation directly in the pixel space by employing the min-max adversarial training. Relatedly, \cite{jahanian2021generative} leverages a pre-trained GAN model to generate the data samples and to replace the entire training dataset like ImageNet~\cite{deng2009imagenet} with a generated one.

\paragraph{View Assimilation }
Previous works like CMC ~\cite{Tian2020ContrastiveMC} and DINO ~\cite{caron2021emerging} study the case of having more than two views in contrastive learning. Both works treat this as a special case in which an arbitrary batch of views has been expanded and demonstrate gains from including additional views. However, empirically we observed that simply augmenting SimCLR with an additional set of positive views degrades its performance. Instead, we explore and propose new methods that can effectively integrate additional views and boost performance.

\section{View Generation}
\subsection{\texorpdfstring{$\mathcal{W}$}{W}-search: perturbation in CL's latent space}
Contrastive learning under the InfoNCE loss leads to congregation of the normalized representations $z / \Vert z \Vert$  of the positive views around their respective anchors on the hypershpere that they reside on. Expectedly, the distribution of the distances of positive views around their anchor is very similar for different anchors. \cref{fig:gan} illustrates this for the case of $z \in \mathbb{R}^3$. Consequently, a straightforward strategy is to generate more positive views from other points in $\mathcal{Z}$ that are the same distance away from the nearest anchor. The main challenge, however, is how to find the corresponding point in the $\mathcal{X}$ space to realize the view?

We propose a simple technique to resolve this challenge by leveraging a pre-trained GAN generator $g$~\cite{karras2020analyzing}. This allows us to pose the view generation as an optimization problem, \ie find a $w^\star$ that minimizes the following loss function. Assuming we are generating $n$ views simultaneously,
\begin{align}
    \{w_k^\star\}_{k=1}^n =& \argmin_{\{w_k\}} \frac{1}{n}\sum_k \underbrace{\delta\left( \epsilon_1, \Vert f \circ g(w_k)-f(x_0) \Vert_2 \right)}_\text{boundary constraint} \nonumber\\
    &+ \underbrace{\lambda (\epsilon_2-\bar{d}_n)^+}_\text{uniformity}. \label{eq:gan}
\end{align}
\noindent Examples of generated views are given in \cref{fig:gan_sample}.

\paragraph{Scalability of $\mathcal{W}$\textit{-search}}
While effective, online $\mathcal{W}$\textit{-search} is computationally expensive because the optimization involving both the contrastive encoder and the generative model needs to be performed for every image in the mini-batch. Therefore, this online view generation using $\mathcal{W}$\textit{-search} does not scale as well to large-scale datasets. This scalability problem can be solved by performing view generation {\em offline}. By leveraging a pretrained $f$, we can cache the generated views before the actual CL training. In the following experiments, we focus on this offline setting via caching, and provide an empirical study of approximated online version in appendix. 

\subsection{\texorpdfstring{$\mathcal{W}$}{W}-perturb: perturbation in GM's latent space}\label{sec:wperturb}
Another solution to the scalability issue of the online $\mathcal{W}$\textit{-search} is to generate views via perturbations in the latent space $\mathcal{W}$ of the generator, instead of the latent space of the contrastive encoder  $\mathcal{Z}$. Thereby, we remove the computationally expensive step of finding views through optimization. 
To avoid on-the-fly optimization, we propose an alternative view generation method, \textit{$\mathcal{W}$\textit{-perturb}}, that creates positive views by directly perturbing in the latent space $\mathcal{W}$ of the pretrained generator $g$.
Under this method, additional positive views for a given anchor $x$ are generated as: 
$\tilde{x} = \mathcal{W}\textit{-perturb}(x) := g(e(x)+w_p)$, where $w_p \sim \mathcal{N}(0, \sigma I)$ and $e(x)$ is the projection of the anchor image in the latent space of $g$. This is a generalization of the {\em latent transforms} $T_\mathbf{z}$ introduced in ~\cite{jahanian2021generative}. The latent transform is not directly applicable to real image domain since the corresponding latents are unknown. Thus, we project real images in GAN's latent space via its inverter $e$.


\section{View Assimilation} \label{sec:view_assim}
In the case of SimCLR, prior works \citep{tamkin2020viewmaker} have explored replacing expert views entirely with generated views. However, a complete replacement leads to a degradation in performance. Taking the MI maximization perspective of the InfoNCE loss \citep{Wu2020OnMI,tian2020makes,van2018representation,poole2019variational} in SimCLR, a possible explanation for the degradation in performance can be attributed to the differences in the MI between the anchor and expert views and the MI between the anchor and generated views. 
We find that while the original and expert views share roughly the same amount of mutual information as the original and generated views, the shared information seem to be different given that there is a similar gap in information between the expert and the generated views. This finding indicates that the generated views are likely to contain meaningfully \textit{complementary} information to the expert views and hence could lead to additional useful features (for the downstream task). Altogether, these observations motivate us to assimilate generated views into contrastive learning, instead of entirely replacing the expert views, to improve downstream accuracy. To this end, we propose two methods for \textit{assimilating} generated views into CL training.

\paragraph{Replacement (A1)}
Our first assimilation method replaces only one of the two expert views with a generated view.  On the generated view, we apply a weak amount of \textit{random-resized-crop} and \textit{flipping}. 

\begin{figure*}[t]
\begin{minipage}[c]{0.3\linewidth}
\centering
\includegraphics[width=0.95\linewidth]{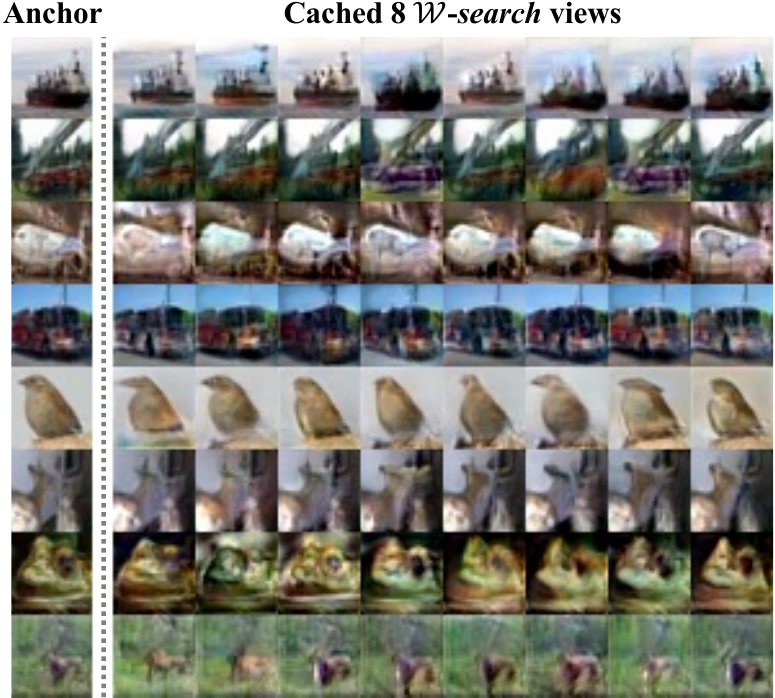}
    \caption{Visual samples of  $\mathcal{W}$\textit{-search} for CIFAR10 dataset. For each row, we show 8 cached views ($\epsilon_1=0.3$).}
    \label{fig:gan_sample}
\end{minipage}\quad
\begin{minipage}[c]{0.68\linewidth}
\centering
\includegraphics[width=0.98\linewidth]{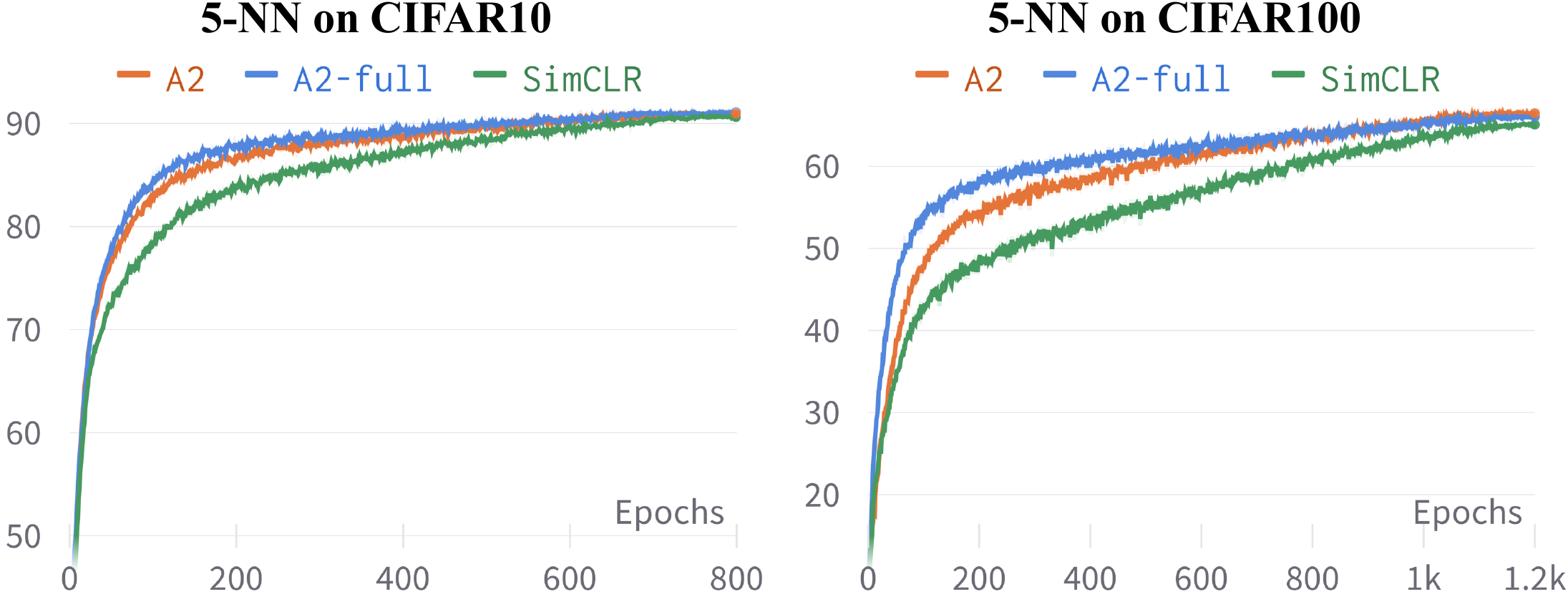}
    \caption{5-NN accuracy curves during training. Models are trained with  $\mathcal{W}$\textit{-search} as the view generation and multiview (A2) as the loss. It is clear that the convergence speed of our models is much faster than that of the baseline SimCLR.}
    \label{fig:converge}
\end{minipage}
\end{figure*}

\paragraph{Multiview (A2)}
Our second assimilation method simply casts the problem as multiview contrastive learning, where there are more than two positive views. We append the additional positive views, $x_{k(i)}$, and define $\{k(i)\}$ as the batch indices of the appended view(s) generated from a anchor image with the index $i$.
This, however, requires an adjustment in the training loss. To this end, we propose the following multiview loss:

\begin{align}
    L_\text{multiview} &= L_\text{CL} - L_\text{align}, \quad \text{where} ~~L_\text{CL}=L_\text{InfoNCE}\nonumber\\
    \text{and}~~L_\text{align} &= \sum_{i \in \mathcal{I}}{\frac{\alpha}{|k(i)|} \sum_{p \in k(i)} z_i^\top z_p / \tau}\label{eq:loss_diag}
\end{align}
\noindent Our loss function appends an $\alpha$-weighted sum of dot products of the projected (to the CL embedding space) anchor view $i$ and its respective generated views $k(i)$ to the base CL loss $L_\text{CL}$. $L_\text{align}$ is a general plug-in term that can be used in conjunction with other existing contrastive losses. However, we empirically found it to work best with InfoNCE ~\citep{van2018representation} and adopt it as our $L_\text{CL}$ unless otherwise specified. Please refer to appendix for results where we use the loss from SimSiam~\citep{chen2021exploring}. We also experimented with the multiview loss from SupCon~\citep{khosla2020supervised} (marked as ``A2-full''), but found it to perform more poorly than our proposed loss. From \cref{fig:converge}, we can see that our models that use our proposed view generation and assimilation strategies exhibit faster convergence than the baseline SimCLR model.

\begin{table*}[t]
    \centering
    \scalebox{0.9}{
    \begin{tabular}{llllcccc}
    \toprule
    View 1 & View 2 & View 3 & Loss & CIFAR10 & CIFAR100 & TinyImageNet & Avg Rank \\ \hline
    expert & expert & \redxmark & SimCLR & 92.04 & 70.41 & 47.48 & 4.67\\
    expert & $\mathcal{W}$\textit{-search} & \redxmark & A1 & 91.86 & 71.69 & {\bf 51.08} & 2.67\\
    expert & $\mathcal{W}$\textit{-perturb} & \redxmark & A1 & 91.09 & 70.83 & 50.18 &  4.67\\
    expert & ViewMaker & \redxmark & A1 & 82.91 & 41.87 & 26.40 & 8.00\\
    ViewMaker & ViewMaker & \redxmark & SimCLR & 83.59 & 44.04 & 40.53 & 7.00\\
    expert & expert & expert & A2 & 91.46 & 70.76 & 47.19 & 5.33\\
    expert & expert & $\mathcal{W}$\textit{-search} & A2 & {\bf 92.90} & \underline{72.76} & \underline{51.05} & {\bf 1.67}\\
    expert & expert & $\mathcal{W}$\textit{-perturb} & A2 & \underline{92.38} & {\bf 72.95} & 50.73 & \underline{2.00}\\
    expert & expert & ViewMaker & A2 & 80.07 & 36.51 & 25.30 & 9.00\\
    \bottomrule
    \end{tabular}}
    \caption{Linear probe accuracy for the four view generation methods ($\mathcal{W}$\textit{-search}, $\mathcal{W}$\textit{-perturb}, Viewmaker~\cite{tamkin2020viewmaker}, expert transformation)  under A1 and A2 view assimilation methods. We also report the baseline SimCLR (reproduced) and Viewmaker (reproduced) accuracies in rows 1 and 5. The highest Acc@1 for each dataset is highlighted in \textbf{boldface}, while the runner-up is \underline{underlined}.}
    \label{tab:main_results}
\end{table*}

\section{Empirical Study}
In this section, we conduct a comprehensive empirical study of view generation and assimilation methods for CL and provide a thorough benchmark of performances. For this purpose, we use the highly optimized SimCLR implementation from \cite{dangovski2021equivariant} for all examined methods and benchmark them on their downstream classification accuracy on four datasets: CIFAR10, CIFAR100 and TinyImageNet. We report the linear probing accuracy as the main evaluation metric. Additional experimental details, hyperparameters, ImageNet experiment and ablations are in appendix.

\subsection{View Generation and Assimilation}
We start with our main results, ablating all possible combinations of four different view generation methods and two view assimilation methods. For the view generation methods, we use our proposed $\mathcal{W}$\textit{-search} and $\mathcal{W}$\textit{-perturb} methods along with Viewmaker ~\citep{tamkin2020viewmaker} and the expert transformations from SimCLR~\citep{chen2020simple}. For assimilation of generated views, we consider replacement of one expert view (A1) and multiview (A2) as described in \cref{sec:view_assim}. For evaluation, we report the top-1 linear probe accuracy (denoted as Acc@1) and $k$-Nearest-Neighbor accuracy ($k=5$, denoted as 5-NN). To obtain linear probe accuracy, we freeze the backbone of $f$ and train a linear layer with SGD for 100 epochs.
To determine the value of $\epsilon$'s for $\mathcal{W}$\textit{-search}, we first pretrain a SimCLR encoder using expert views and compute the $\epsilon$ as the average distance (in $\mathcal{W}$-space) between anchors and their expertly transformed views. For $\mathcal{W}$\textit{-perturb}, we conduct a grid search on the $\sigma$. 

As shown in \cref{tab:main_results}, both $\mathcal{W}$\textit{-search} and $\mathcal{W}$\textit{-perturb} outperform all other view generation methods on CIFAR10, CIFAR100, and TinyImageNet. When we replace one of the views with our proposed view generation strategies, except in CIFAR10, we see consistent improvements and $\mathcal{W}$\textit{-search} proves to be a more effective generation method. Especially for TinyImageNet, we see an improvement of $\approx 3.6\%$. When we augment the generated views for multiview contrastive learning, in contrary to the intuition that more expert views should improve performance, assimilating a third expert view in fact degrades performance in most cases. On the other hand, the views we generate with $\mathcal{W}$\textit{-search} and $\mathcal{W}$\textit{-perturb} consistently lead to improvements of $0.9\%, 2.3\%,$ and $3.6\%$ on CIFAR10, CIFAR100 and TinyImageNet respectively. 
Overall, our generated views, when replacing one view or being assimilated, almost always leads to an improvement suggests that our framework for view generation allows for generating views that capture some different information from the expert transformations for the downstream task.

\section{Conclusion and Limitations}
In this work, we presented an empirical study on the view generation and view assimilation techniques in contrastive learning. We showed that when used in conjunction with expert-views, generated views consistently improve downstream classification performance on three different datasets. However, the improvement in the performance is contingent on not only the method of view-generation but more heavily on how the generated view is assimilated, which has been not explored before to our knowledge.

\clearpage
{\small
\bibliographystyle{ieee_fullname}
\bibliography{ref}
}

\clearpage
\appendix

\section*{Appendix}

\section{Experimental Setup and Details} \label{app:exp}
\noindent \textbf{Datasets.} Experiments are conducted on four datasets: 
\begin{itemize}
    \item CIFAR10~\citep{krizhevsky2009learning} has 10 classes, 50,000 images for training and 10,000 for testing.
    \item CIFAR100~\citep{krizhevsky2009learning} has 100 classes, 50,000 images for training and 10,000 for testing.
    \item TinyImageNet is introduced in \cite{le2015tiny}. The dataset contains 200 classes, 100,000 images for training and 10,000 for testing. Images are resized to $64 \times 64$.
    \item ImageNet~\citep{deng2009imagenet} contains approximately 1.3 million images. We following the setting in \cite{jahanian2021generative} and generate 1.3 million ``fake'' images using a pretrained BigBiGAN~\citep{donahue2019large} at resolution $128 \times 128$.
\end{itemize}

\noindent \textbf{Implementation.} We implement our methods based on the E-SSL~\cite{dangovski2021equivariant} codebase\footnote{{https://github.com/rdangovs/essl/tree/main/cifar10}} (for CIFAR10 experiments) and the SupCon~\cite{khosla2020supervised} codebase\footnote{{https://github.com/HobbitLong/SupContrast}} (for CIFAR100 and TinyImageNet). Hyperparameters for each dataset are listed in \cref{tab:exp_hyperparam}\footnote{For CIFAR10, we found that batch size of 128 gives similar or slightly better results than the default 512.}.

\noindent \textbf{Viewmaker.} For Viewmaker~\cite{tamkin2020viewmaker}, we reproduced the reported accuracy on CIFAR10. For CIFAR100 and TinyImageNet, since the original authors did not evaluate their model against these datasets, we tried our best to optimize the hyperparameters, such as optimizer, learning rate, temperature, architecture of encoder (ResNet18 small, ResNet18, ResNet50), and projection head. We describe the best set of hyperparameters in \cref{tab:viewmaker_hyperparam}.

\noindent \textbf{Evaluation.}
As for evaluation metrics, we adopt the conventions in the respective codebases. For CIFAR10, we run linear probe evaluations (for 100 epochs) with 5 random seeds and report the mean and standard deviation of accuracies. For CIFAR100 and TinyImageNet, we run linear probe evaluations for 100 epochs and report the best accuracy. For both settings, we load and freeze the last checkpoint of the backbone network.
\begin{table}[h]
    \centering
    \resizebox{1\linewidth}{!}{
    \begin{tabular}{lccc}
    \toprule
                  & CIFAR10 & CIFAR100 & TinyImageNet \\\hline
    Optimizer     & SGD     & SGD      & SGD \\
    Learning Rate & 0.015   & 0.5      & 0.5 \\
    Weight Decay  & 5e-5    & 1e-4     & 1e-4 \\
    Momentum      & 0.9     & 0.9      & 0.9 \\
    Cosine Decay  & \greencmark & \greencmark & \greencmark \\
    Batch Size    & 128     & 512      & 512 \\
    SimCLR Loss   & InfoNCE & SimCLR & SimCLR \\
    Temperature   & 0.5     & 0.5      & 0.5 \\
    Epochs        & 800     & 1200     & 1000 \\
    Backbone      & ResNet18 & ResNet50 & ResNet18 \\
    Embedding Dim & 512    & 2048      & 512 \\
    Projection Dim & 2048     & 128      & 128 \\
    \bottomrule
    \end{tabular}}
    \caption{Hyperparameters for experiments.}
    \label{tab:exp_hyperparam}
\end{table}
\begin{table}[h]
    \centering
    \resizebox{1\linewidth}{!}{
    \begin{tabular}{lccc}
    \toprule
                  & CIFAR10 & CIFAR100 & TinyImageNet \\\hline
    Optimizer     & \multicolumn{3}{c}{SGD (for encoder), Adam (for Viewmaker module)} \\
    Learning Rate & 0.015   & 0.06      & 0.06 \\
    Weight Decay  & 1e-4    & 1e-4     & 1e-4 \\
    Momentum      & 0.9     & 0.9      & 0.9 \\
    Cosine Decay  & \redxmark & \redxmark & \redxmark \\
    Batch Size    & 128     & 512      & 128 \\
    SimCLR Loss   & ViewMaker & ViewMaker & ViewMaker \\
    Temperature   & 0.07     & 0.1      & 0.5 \\
    $\alpha$ for A2 loss    & 0.14 & 0.1 & 0.5\\
    Epochs        & 200     & 800     & 800 \\
    Backbone      & ResNet18 & ResNet50 & ResNet18 \\
    Noise Dim & 100     & 100       & 100 \\
    Embedding Dim & 512    & 2048      & 512 \\
    Projection Dim & 128     & 128      & 128 \\
    \bottomrule
    \end{tabular}}
    \caption{Hyperparameters for our reproduced ViewMaker~\cite{tamkin2020viewmaker} experiments.}
    \label{tab:viewmaker_hyperparam}
\end{table}

\noindent \textbf{Computational cost for view generation.}
The computation time depends on the hyperparameters (the number of optimization steps for $\mathcal{W}$\textit{-search}), \eg, caching 8 views per sample for CIFAR10 takes 12.19 A100 GPU hours.

\noindent \textbf{Computational cost for pretraining.} Each experiment is run on 4 NVIDIA V100 GPUs. The pretraining time of SimCLR baseline for CIFAR10, CIFAR100, and TinyImagenet are 11.5, 13.6, and 29.7 hours, respectively.

\section{Mutual Information Analysis}
Taking the mutual information (MI) maximization perspective of the InfoNCE \cite{Wu2020OnMI,tian2020makes,van2018representation,poole2019variational} loss in SimCLR, \cite{Tian2020ContrastiveMC, tian2020makes} introduced an information theoretic definition of what makes for a ``good'' positive view in contrastive learning. While it is difficult to accurately estimate MI in the high-dimensional $\mathcal{X}$ space, in Table~\ref{tab:mi_c110}, we provide rough estimates for initial analysis. We find that the original and generated views share similar or lower amount of mutual information than the original and expert views. However, generated ($\mathcal{W}$\textit{-search} or $\mathcal{W}$\textit{-perturb}) and expert views share even lower mutual information than original and expert views, which indicates that VG1 views are likely to contain meaningfully complementary information to the expert views and hence lead to additional useful features (for the downstream task). These observations motivate us to assimilate generated views into contrastive learning, instead of entirely replacing the expert views, to improve downstream accuracy.

\begin{table}[h]
    \centering
    \scalebox{1}{
    \begin{tabular}{lcc}
    \toprule
    View Pairs & {CIFAR10} & {CIFAR100} \\\hline
    Original, Expert & 4.14 & 5.41 \\
    Original, $\mathcal{W}$\textit{-search} & 4.13 & 4.40 \\
    $\mathcal{W}$\textit{-search}, Expert  & 3.78 & 4.35 \\
    Original, $\mathcal{W}$\textit{-perturb} & 3.79 & 3.91 \\
    $\mathcal{W}$\textit{-perturb}, Expert  & 3.66 & 3.86 \\
    \bottomrule
    \end{tabular}}
    \caption{Estimated mutual information between the anchor image (original), positive expert view (expert) and generated views ($\mathcal{W}$\textit{-search} and $\mathcal{W}$\textit{-perturb}) using the MINE~\cite{belghazi2018mine} estimator.}
    \label{tab:mi_c110}
\end{table}

\section{Details of StyleGAN and In-Domain GAN Inversion}
\paragraph{StyleGAN}
We use StyleGAN2\footnote{{https://github.com/rosinality/stylegan2-pytorch}} for our experiments on CIFAR and TinyImageNet. The StyleGAN generator consists of two key components: (1) a mapping function $g_1: \mathcal{S} \mapsto \mathcal{W}$ that maps the Gaussian-distributed latent code $s \in \mathcal{S}$ into a collection of style codes $w \in \mathcal{W}$, and (2) a generator $g: \mathcal{W} \mapsto \mathcal{X}$ that decodes $w \in \mathcal{W}$ to an image. Here, $w$ is a concatenation of $w_1, ..., w_k$, where each $w_i$ corresponds to the style code from the $i^\text{th}$ convolutional block of $g$.
\paragraph{In-Domain GAN Inversion}
Let $e: \mathcal{X} \mapsto \mathcal{W}$ denote an inverter neural network. Let $d: \mathcal{X} \mapsto \mathbb{R}$ denote the discriminator network. In-domain GAN inversion~\cite{zhu2020domain} aims to learn a mapping from images to latent space. The encoder is trained to reconstruct real images (thus are ``in-domain'') and guided by image-level loss terms, \ie pixel MSE, VGG perceptual loss, and discriminator loss:
\begin{align}
    L_\text{idinv}(e,d,g) = &\mathbb{E}_{x \sim P_{X}} [ \Vert x-g \circ e(x) \Vert_2 +\nonumber\\
    &\lambda_\text{vgg} \Vert h(x)-h \circ g \circ e(x) \Vert_2 -\nonumber\\
    &\lambda_\text{adv} \mathcal{A}({-\tilde{d}\circ g \circ e(x)}) ], \label{eq:idinv}
\end{align}
\noindent where $h$ is perception network and here we keep the same as in-domain inversion as VGG network, ${a}$ is the {\em activation function} and $\tilde{d}$ is the {\em logit} or discriminator's output before activation. Note that choosing ${a}(t)=\text{softplus}(t)=\log{(1+\exp{(t)})}$ recovers the original GAN formulation~\cite{goodfellow2014generative,karras2019style}, and the resulting objective minimizes the Jensen-Shannon divergence between real and generated data distributions. After encoder training, we optimize the associated latent variable $w$ for each image $x$ with the same loss function using $w=e(x)$ as a warm start. Note that in the main text we reload the notation $e(\cdot)$ as the final results after $w$-optimization, which are precomputed and cached.

\section{Details of Training Loss}
\begin{figure*}[t]
    \centering
    \includegraphics[width=0.8\linewidth]{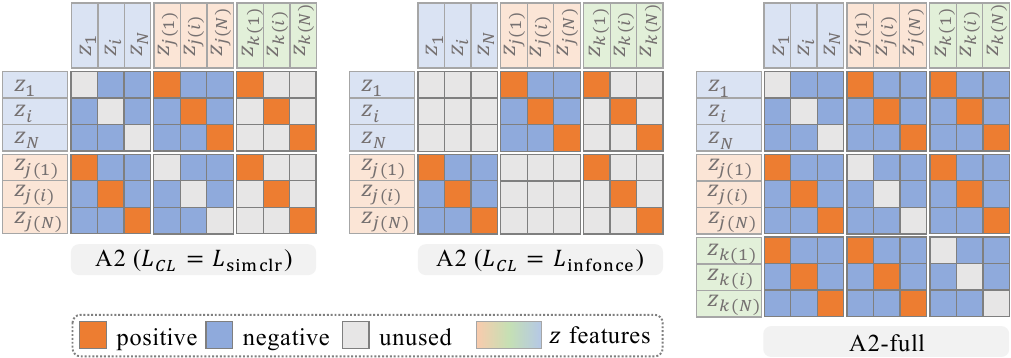}
    \caption{Visual illustration of the three contrastive loss functions: A2-SimCLR, A2-InfoNCE, and A2-full for a mini-batch size of $N$ images (visualizing $N=3$). Each small square represents the inner-product between corresponding features. The contrastive loss for each instance $i$ is defined on the corresponding row of inner-products. }
    \label{fig:contra_loss}
\end{figure*}
We provide an illustrative visualization of our A2 losses in \cref{fig:contra_loss}. The detailed formulation of loss functions are as follows,
\begin{align}
    L_{\text{infonce}} =& -\sum_{i \in \mathcal{I}_1}{\log{ \frac{\exp(z_i \cdot z_{j(i)}/\tau)}{\sum_{a \in \mathcal{I}_2} \exp(z_i \cdot z_{a}/\tau)} } } \nonumber\\
    &-\sum_{i \in \mathcal{I}_2}{\log{ \frac{\exp(z_i \cdot z_{j(i)}/\tau)}{\sum_{a \in \mathcal{I}_1} \exp(z_i \cdot z_{a}/\tau)} } } \label{eq:infonce}\\
    L_\text{A2\text{-}\text{full}} =& -\sum_{i \in \mathcal{I}}{\frac{-1}{|P(i)|}\sum_{p \in P(i)}\log{\frac{\exp(z_i \cdot z_p /\tau)}{\sum_{a \in \mathcal{A}(i)} \exp(z_i \cdot z_{a}/\tau)}}} \nonumber\\
    \text{where}&~P(i) = \{j(i)\} \cup \{k(i)\}\label{eq:loss_supcon}\\
    L_\text{A2-simclr} &= L_\text{simclr} - \sum_{i \in \mathcal{I}}{\frac{\alpha}{|k(i)|} \sum_{p \in k(i)} z_i \cdot z_p / \tau} \nonumber\\
    = -\sum_{i \in \mathcal{I}}&\log{ \frac{\exp(z_i \cdot z_{j(i)} /\tau + \frac{\alpha}{|k(i)|}\sum_{p \in k(i)} (z_i \cdot z_{p})/\tau)}{\sum_{a \in \mathcal{A}(i)} \exp(z_i \cdot z_{a}/\tau)} }\label{eq:loss_diag}
\end{align}
In $L_\text{infonce}$, $\mathcal{I}_1$ and $\mathcal{I}_2$ are the set of indices of two positive views. An ablation of A2 losses is provided in \cref{tab:ablation_loss_c110}. We observe that A2-InfoNCE performs the best for both datasets. We used A2-InfoNCE as our A2 loss if not specified.
\begin{table}[h]
    \centering
    \resizebox{1\linewidth}{!}{
    \begin{tabular}{lcccc}
    \toprule
     & \multicolumn{2}{c}{CIFAR10} & \multicolumn{2}{c}{CIFAR100} \\
    \cmidrule(lr){2-3} \cmidrule(lr){4-5}
    Loss & Acc@1  & 5-NN & Acc@1 & 5-NN \\\hline
    A2-full      & 92.57 & 91.57 & 71.82 & 66.06 \\
    A2-SimCLR ~($\alpha=0.5$) & 92.66 & 91.05 & 72.27 & 65.85 \\
    A2-InfoNCE ~($\alpha=0.5$) & \textbf{92.90} & 90.95 & \textbf{72.76} & 66.46 \\
    A2-InfoNCE ~($\alpha=1$)   & 92.54 & 90.80 & 72.61 & 66.96 \\
    \bottomrule
    \end{tabular}}
    \caption{Ablation on the loss functions with VG1 views on the CIFAR10 and CIFAR100 datasets. A2-SimCLR is $L_\text{simclr} - L_\text{align}$ and A2-InfoNCE is $L_\text{infonce} - L_\text{align}$.}
    \label{tab:ablation_loss_c110}
\end{table}

\section{SimSiam Experiments}\label{sec:simsiam}
\begin{figure}[t]
    \centering
    \includegraphics[width=0.95\linewidth]{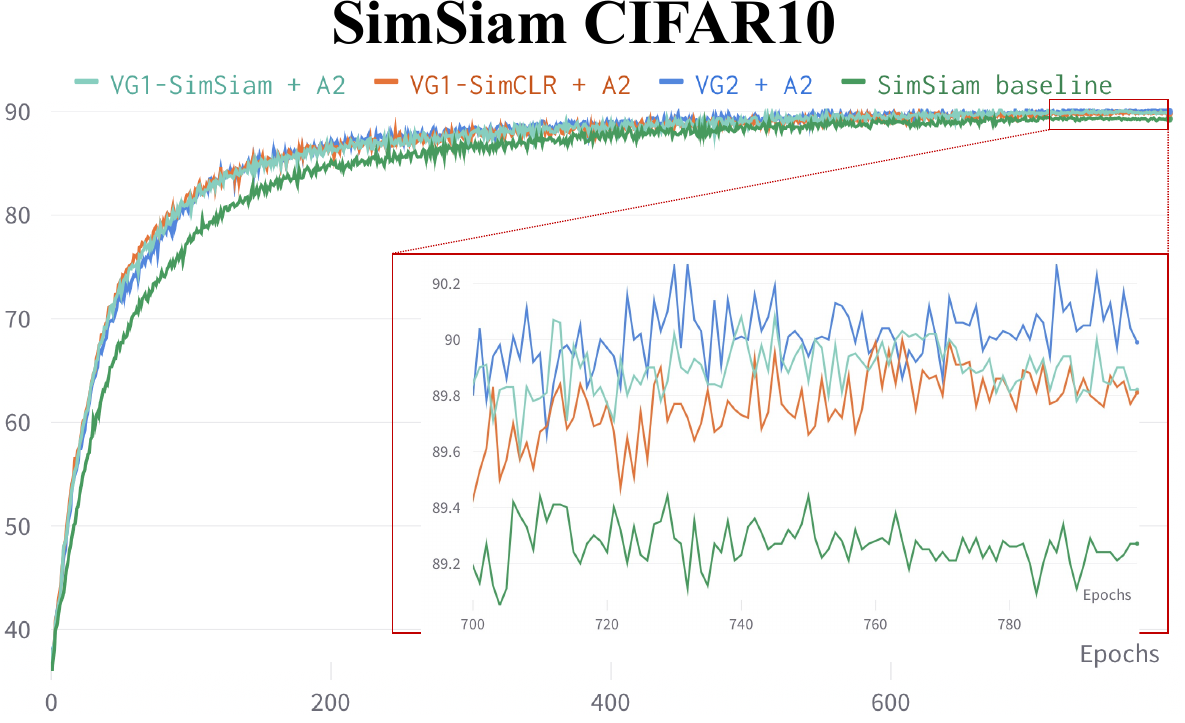}
    \caption{5-NN accuracy curves of SimSiam experiments on CIFAR10. In legend, VG1 refers to $\mathcal{W}$\textit{-search} and VG2 refers to $\mathcal{W}$\textit{-perturb}.}
    \label{fig:simsiam}
\end{figure}

We conducted experiments of SimSiam~\cite{chen2021exploring} with generated views. The 5-NN accuracy curves are reported in \cref{fig:simsiam}. The linear probe accuracy is reported in \cref{tab:simsiam}. The A2 loss needs to be adjusted accordingly to avoid training collapse,
\begin{align}
    &L_\text{A2-simsiam} = L_\text{simsiam} \nonumber\\
    & +\sum_{i \in I}{\frac{\alpha}{|k(i)|} \sum_{p \in k(i)} D(\text{predictor}(z_p), \text{stopgrad}(z_i)) }\label{eq:a2_simsiam}
\end{align}
\noindent where $D$ is the cosine similarity and $L_\text{simsiam}$ is the SimSiam loss function.
\begin{table}[t]
    \centering
    \scalebox{1}{
    \begin{tabular}{lcc}
    \toprule
     & Acc@1 & 5-NN \\ \hline
    SimSiam baseline & 90.17 & 89.27 \\
    $\mathcal{W}$\textit{-search}-SimCLR + A2-SimSiam  & \textbf{90.96} & 89.81 \\
    $\mathcal{W}$\textit{-search}-SimSiam + A2-SimSiam & 90.79 & 89.82 \\
    $\mathcal{W}$\textit{-perturb} + A2-SimSiam         & 90.28 & \textbf{89.99} \\
    \hline
    SimCLR baseline          & 92.04 & 90.65 \\
    $\mathcal{W}$\textit{-search}-SimSiam + A2-InfoNCE & \textbf{92.44} & 90.88 \\
    \bottomrule
    \end{tabular}}
    \caption{Linear-probing and 5-NN accuracies of SimSiam experiments on CIFAR10.}
    \label{tab:simsiam}
\end{table}

\section{Online \texorpdfstring{$\mathcal{W}$}{W}-search}\label{sec:vg1_online}
\begin{figure}[t]
    \centering
    \includegraphics[width=0.95\linewidth]{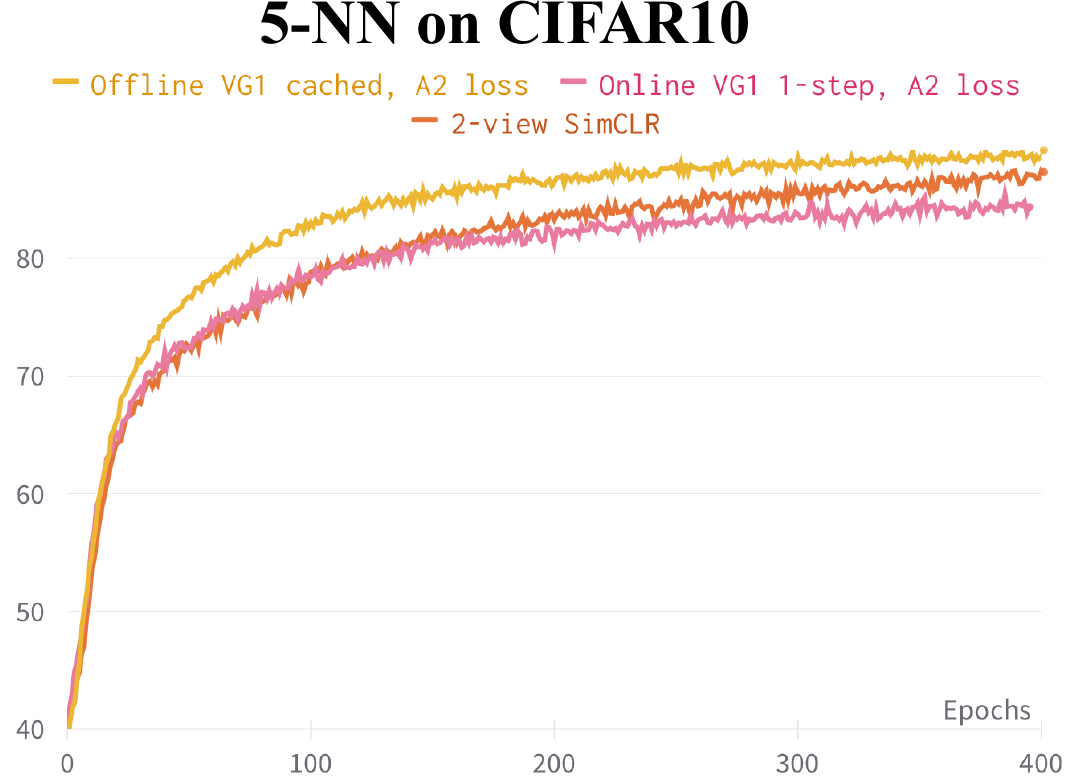}
    \caption{Online $\mathcal{W}$\textit{-search}. In legend, VG1 refers to $\mathcal{W}$\textit{-search}.}
    \label{fig:vg1_online}
\end{figure}

In the online $\mathcal{Z}$-search setting, the optimization is performed involving the current SimCLR encoder during training. We tried to perform 1-step optimization with fast sign gradient~\cite{goodfellow2014explaining}, but observed the results are worse than the SimCLR baseline. The 5-NN accuracies are reported in \cref{fig:vg1_online}.

\section{Ablation on Hyperparameters} \label{app:hyperparams}
In this section we provide ablations on hyperparameters $\epsilon_1$, $\epsilon_2$, and $\lambda$ introduced in $\mathcal{Z}$-search. In addition, we perform grid search on $\sigma$ introduced in $\mathcal{Z}$-perturb.

\noindent \textbf{Ablation on $\epsilon_1$ and $\epsilon_2$.}
We conduct ablation studies of $\epsilon_1$, $\epsilon_2$, and $\lambda$ on CIFAR10. In \cref{tab:ablation_eps12}, we set $\epsilon_2=\epsilon_1+0.2$ except for $\epsilon_1$ of values 0.1 and 0.2. We empirically find that it is difficult to reach a large pairwise distance $\epsilon_2$ when $\epsilon_1$ is small, and a large $\epsilon_2$ leads to more optimization steps. By design, a large $\epsilon_2$ encourages generating diverse samples. A {\em rule of thumb} is to set $\epsilon_2 \geq \epsilon_1$.

\noindent \textbf{Ablation on $\lambda$.}
In \cref{tab:ablation_eps1}, we fix $\epsilon_2=0.15$ and vary $\epsilon_1$. In \cref{tab:ablation_lam}, we fix $\epsilon_1=0.3$ and $\epsilon_2=0.5$ and vary $\lambda$.

\noindent \textbf{Ablation on $\sigma$.}
We perform grid search on $\sigma$ for VG2 and report results in \cref{tab:ablation_gauss_c10} and \cref{tab:ablation_gauss_c100}. We find that for both CIFAR10 and CIFAR100, $\sigma=0.2$ leads to the best results, which is consistent with the empirical findings in ~\cite{jahanian2021generative}.
\begin{table}[t]
    \centering
    \scalebox{1}{
    \begin{tabular}{lc}
    \toprule
    View & Acc@1 \\
    \hline
    $e(x)$                  & {91.697} $\pm$ 0.038 \\
    $e(x)+w_\texttt{Gauss} \sim N(0, 0.1)$ & {92.297} $\pm$ 0.024 \\
    $e(x)+w_\texttt{Gauss} \sim N(0, 0.2)$ & {\bf 92.383} $\pm$ 0.033 \\
    $e(x)+w_\texttt{Gauss} \sim N(0, 0.4)$ & {91.852} $\pm$ 0.030 \\
    $e(x)+w_\texttt{Gauss} \sim N(0, 1.0)$ & {87.893} $\pm$ 0.026 \\
    \bottomrule
    \end{tabular}}
    \caption{Ablation on $\sigma$ of value 0, 0.1, 0.2, 0.4 and 1.0. Experiments are conducted on CIFAR10 with A2 loss.}
    \label{tab:ablation_gauss_c10}
\end{table}
\begin{table}[t]
    \centering
    \scalebox{1}{
    \begin{tabular}{lcc}
    \toprule
    View & Acc@1 & 5-NN \\ \hline
    $e(x)+w_\texttt{Gauss} \sim N(0, 0.1)$ & 71.84 & 66.33 \\
    $e(x)+w_\texttt{Gauss} \sim N(0, 0.2)$ & {\bf 72.95} & {\bf 66.60} \\
    $e(x)+w_\texttt{Gauss} \sim N(0, 0.4)$ & 71.69 & 65.33 \\
    $e(x)+w_\texttt{Gauss} \sim N(0, 1.0)$ & 68.05 & 61.06 \\
    \bottomrule
    \end{tabular}}
    \caption{Ablation on $\sigma$ of value 0.1, 0.2, 0.4 and 1.0. Experiments are conducted on CIFAR100 with A2 loss.}
    \label{tab:ablation_gauss_c100}
\end{table}
\begin{table}[h]
    \centering
    \scalebox{1}{
    \begin{tabular}{llc}
    \toprule
    $\epsilon_1$ & $\epsilon_2$ & Acc@1 \\
    \hline
    0.1      & 0.15 & 92.584 $\pm$ 0.023 \\
    0.2      & 0.35 & 92.615 $\pm$ 0.048 \\
    0.3      & 0.50 & {\bf 92.898} $\pm$ 0.045 \\
    0.5      & 0.70 & 92.451 $\pm$ 0.053 \\
    0.7      & 0.90 & 91.760 $\pm$ 0.032 \\
    0.9      & 1.10 & 91.561 $\pm$ 0.051 \\
    \bottomrule
    \end{tabular}}
    \caption{Ablation on $\epsilon_1$ and $\epsilon_2$, $\lambda=0.01$. Experiments are conducted on CIFAR10.}
    \label{tab:ablation_eps12}
\end{table}
\begin{table}[h]
    \centering
    \scalebox{1}{
    \begin{tabular}{llc}
    \toprule
    $\epsilon_1$ & $\epsilon_2$ & Acc@1 \\
    \hline
    0.1      & 0.15 & 92.584 $\pm$ 0.023 \\
    0.2      & 0.15 & {92.385} $\pm$ 0.043 \\
    0.3      & 0.15 & \textbf{92.643} $\pm$ 0.043 \\
    0.5      & 0.15 & {92.455} $\pm$ 0.032 \\
    0.7      & 0.15 & {92.174} $\pm$ 0.029 \\
    \bottomrule
    \end{tabular}}
    \caption{Ablation on $\epsilon_1$, $\epsilon_2=0.15$ and $\lambda=0.01$. Experiments are conducted on CIFAR10 with the A2 loss.}
    \label{tab:ablation_eps1}
\end{table}
\begin{table}[h]
    \centering
    \scalebox{1}{
    \begin{tabular}{llc}
    \toprule
    $\epsilon_1$ & $\lambda$ & Acc@1 \\
    \hline
    0.3      & 0     & {92.555} $\pm$ 0.028 \\
    0.3      & 0.005 & {92.756} $\pm$ 0.042 \\
    0.3      & 0.01 & {\bf 92.898} $\pm$ 0.045 \\
    0.3      & 0.02 & {92.854} $\pm$ 0.027 \\
    \bottomrule
    \end{tabular}}
    \caption{Ablation studies on $\lambda$. For all entries we fix $\epsilon_1=0.3$ and $\epsilon_2=0.5$. Experiments are conducted on CIFAR10 with the A2 loss.}
    \label{tab:ablation_lam}
\end{table}
\begin{table}[h]
    \centering
    \scalebox{1.0}{
    \begin{tabular}{lllll}
    \toprule
    View 1 & View 2 & View 3 & Loss & Acc@1 \\ \hline
    expert & expert & \redxmark & SimCLR & 49.93 \\
    expert & $\mathcal{W}$\textit{-perturb} & \redxmark & A1 & 48.81 \\
    expert & expert & $\mathcal{W}$\textit{-perturb} & A2 & {\bf 51.42} \\
    \bottomrule
    \end{tabular}}
    \caption{ImageNet experiments. Linear probe accuracies are reported.}
    \label{tab:imagenet}
\end{table}

\section{ImageNet Experiments}
We also evaluate against the standard large-scale dataset ImageNet. Following the more difficult setting in \cite{jahanian2021generative}, we generate 1.3 million ``fake'' images using BigBiGAN~\citep{donahue2019large} (approximately the number of images in ImageNet) from the GAN and only train on this generated dataset. However, we follow the standard protocol of reporting the linear probe accuracy on the real ImageNet dataset. As evident in \cref{tab:imagenet}, adding our generated view as an additional positive view improves performance by $\approx 1.5\%$ and proves to capture some meaningful information that expert transformations do not.

\end{document}